%% file: main.tex
\documentclass[lettersize,journal]{IEEEtran}
\usepackage{amsmath,amsfonts}
\usepackage{algorithmic}
\usepackage{array}
\usepackage[caption=false,font=normalsize,labelfont=sf,textfont=sf]{subfig}
\usepackage{textcomp}
\usepackage{stfloats}
\usepackage{url}
\usepackage{verbatim}
\usepackage{graphicx}
\usepackage{algorithm}
\usepackage{algorithmic}
\usepackage{color}
\usepackage{soul}
\def\BibTeX{{\rm B\kern-.05em{\sc i\kern-.025em b}\kern-.08em
    T\kern-.1667em\lower.7ex\hbox{E}\kern-.125emX}}
\usepackage{balance}
\begin{document}

\title{Hierarchical Dual-Strategy Unlearning for Biomedical and Healthcare Intelligence Using Imperfect and Privacy-Sensitive Medical Data}

\author{Yi~Zhang$^{*}$,
        Chao~Zhang$^{*}$,
        Zijian~Li,
        Tianxiang~Xu,
        Kunyu~Zhang,
        Zhan~Gao,
        Meinuo~Li,
        Xiaohan~Zhang,
        Qichao~Qi$^{\dagger}$,
        and~Bing~Chen$^{\dagger}$
\thanks{$^{*}$These authors contributed equally to this work.}
\thanks{$^{\dagger}$Corresponding authors.}
\thanks{C. Zhang and Q. Qi are with the Department of Neurosurgery, Qilu Hospital of Shandong University and Institute of Brain and Brain-Inspired Science, Jinan 250012, China (e-mail: qiqichao@sdu.edu.cn).}
\thanks{B. Chen is with the Department of Neurosurgery, The Affiliated Hospital of Qingdao University, Qingdao, Shandong 266000, China (e-mail: chenbing\_sjwk@qduhospital.cn).}
\thanks{Y. Zhang is with the Department of Neurosurgery, Peking Union Medical College Hospital, Beijing 100730, China.}
\thanks{T. Xu is with the School of Software and Microelectronics, Peking University, Beijing 102600, China.}
\thanks{Z. Li is with the College of Artificial Intelligence, Dalian Maritime University, Dalian 116026, China.}
\thanks{K. Zhang is with the University of Colorado Boulder, Boulder, CO 80309, USA.}
\thanks{Z. Gao is with Zhengzhou University, Zhengzhou, Henan 450001, China.}
\thanks{M. Li is with The University of Hong Kong, Hong Kong SAR, China.}
\thanks{X. Zhang is with the Georgia Institute of Technology, Atlanta, GA 30332, USA.}}

\maketitle

\begin{abstract}
Large language models (LLMs) exhibit exceptional performance across diverse domains, yet their propensity to memorise training data poses substantial privacy risks, particularly within sensitive healthcare contexts where medical data is often imperfect, insufficiently labelled, or contains privacy-sensitive patient information. Here, we present a hierarchical dual-strategy framework for selective knowledge unlearning in medical LLMs that enables precise removal of specialised knowledge whilst preserving fundamental medical competencies, specifically addressing challenges in biomedical and healthcare intelligence using imperfect data. Our approach synergistically integrates geometric-constrained gradient updates, which selectively modulate parameters encoding target knowledge while safeguarding essential capabilities, with concept-aware token-level interventions that systematically distinguish between preservation-critical and unlearning-targeted tokens through a unified four-level medical concept hierarchy. Through comprehensive evaluation on the MedMCQA dataset targeting surgical knowledge removal, and cross-domain validation on the MHQA dataset encompassing anxiety, depression, trauma, and obsessive-compulsive disorder domains, we demonstrate that our method achieves superior selective unlearning performance (82.7\% forgetting rate, 88.5\% knowledge preservation) compared to existing approaches. Notably, our framework maintains robust privacy guarantees while requiring modification of only 0.1\% of model parameters, establishing a paradigm for privacy-preserving medical AI systems that addresses regulatory compliance, hospital auditability, and ethical imperatives in clinical research environments. This work contributes to advancing biomedical and healthcare intelligence by providing an effective solution for auditing and managing imperfect and privacy-sensitive medical data in real-world clinical applications.
\end{abstract}

\begin{IEEEkeywords}
machine unlearning, imperfect medical data, privacy-sensitive healthcare data, biomedical intelligence, selective knowledge removal, hierarchical medical concepts, weakly supervised learning
\end{IEEEkeywords}

\input{body/introduction}

\input{body/related_work}

\input{body/methodology}

\input{body/experiments}

\input{body/conclusion}

\end{document}

%% file: body/introduction.tex
\section{Introduction}

Large language models (LLMs) have transformed healthcare informatics, demonstrating remarkable capabilities in medical question-answering and clinical decision support. However, their deployment faces significant challenges when dealing with imperfect medical data, which is characteristically incomplete, insufficiently labelled, imbalanced, or contains annotation noise \cite{zhang2025clinicalexpertuncertaintyguided}. Moreover, their ability to memorize training data raises substantial privacy concerns when deployed on sensitive medical datasets. Critically, current methodologies lack the capability to selectively excise specific sensitive information from imperfect, interconnected medical datasets without compromising the model's broader clinical reasoning~\cite{han2025llmwebshell}. Privacy regulations such as GDPR emphasize the "right to be forgotten," necessitating robust machine unlearning methodologies that can effectively manage imperfect and privacy-sensitive medical data for responsible AI deployment \cite{Fan2024SalUnEM, huang2024unified}.

Medical domain unlearning faces critical challenges when dealing with imperfect healthcare data. LLMs may inadvertently encode patient-specific information from insufficiently anonymised data, creating privacy risks \cite{chen2024robust}. Rapidly evolving medical guidelines require models to "forget" outdated or incorrectly labelled information \cite{khan2024med42}. Specialized medical knowledge trained on imbalanced datasets requires compartmentalization for selective access \cite{xiong2023knowledge}. 
For instance, a compliant clinical AI system must retain general diagnostic capabilities (e.g., identifying common symptoms of brain tumors) while selectively unlearning restricted surgical procedural details (e.g., specific steps for brain tumor resection) to ensure patient safety and regulatory adherence.
This is particularly important for mental health specialties, where conditions such as anxiety, depression, trauma, and obsessive-compulsive disorders demand heightened privacy protection whilst dealing with often incomplete or noisy diagnostic data. Hospital research environments require frameworks that selectively manage imperfect and sensitive information whilst preserving clinical utility. Advanced attention mechanisms and multi-scale analysis techniques from computer vision~\cite{zhang2024cf, zhang2023multi} have inspired similar approaches in medical data processing to improve model robustness.

Traditional unlearning approaches face significant challenges when applied to imperfect medical data. Knowledge boundary delineation is complex due to interconnected medical domains and incomplete supervision, surgical knowledge shares fundamental concepts with other specialities whilst dealing with varying annotation quality \cite{geng2025comprehensive}. Knowledge integrity preservation requires understanding hierarchical medical concept organisation in the presence of label noise and data imbalance \cite{wang2025kgt}. Imperfect medical data imposes stringent privacy requirements demanding rigorous removal guarantees whilst maintaining model utility on insufficiently labelled datasets \cite{yu2022differentially}.

Existing methods include complete retraining (strong guarantees but computationally prohibitive \cite{huang2024unified}) and gradient-based approaches (efficient but limited precision on noisy data \cite{liu2024unlearning}). Recent advances in multimodal and token-level unlearning show promise \cite{huo2025mmunlearner, tran2025tokens} but haven't adequately addressed the specific challenges of managing imperfect medical data with incomplete supervision and privacy constraints.

We introduce a hierarchical dual-strategy framework combining geometric-constrained gradient updates with concept-aware token interventions through a unified four-level medical concept hierarchy (L1: fundamental biomedical, L2: general clinical, L3: specialty-specific, L4: surgical concepts). The geometric component uses Fisher Information Matrix analysis to selectively modify surgical parameters while preserving general medical reasoning \cite{yao2024sugd}. The token component employs gradient-based importance scoring to identify surgical tokens while maintaining fundamental medical vocabulary \cite{jang2022knowledge}.

We evaluate on MedMCQA (surgical unlearning) and MHQA datasets \cite{joshiprashanthd2025mhqa} (mental health domains: anxiety, depression, trauma, OCD). Our approach demonstrates superior selective unlearning performance, outperforming existing methods with robust privacy guarantees while requiring minimal parameter modification \cite{huang2023tight, icml2024dora}.

Key contributions include: 

\begin{itemize}
    \item A hierarchical dual-strategy framework addressing unlearning at parameter and vocabulary levels, specifically designed for imperfect medical data management;
    \item A hierarchical medical concept methodology for precise targeting whilst handling incomplete supervision and annotation noise;
    \item A comprehensive evaluation framework assessing effectiveness, preservation, privacy, and efficiency on real-world imperfect medical datasets;
    \item Empirical evidence demonstrating superiority in biomedical and healthcare intelligence using imperfect data.
\end{itemize}

This establishes a paradigm for privacy-preserving medical AI addressing regulatory and ethical requirements whilst effectively managing imperfect healthcare data \cite{li2023dp}.

%% file: body/related_work.tex
\section{Related Work}

\subsection{Machine Unlearning}

Machine unlearning removes specific knowledge from trained models to address privacy regulations and ethical considerations. Exact methods like complete retraining \cite{huang2024unified} provide strong guarantees but are computationally expensive. Approximate methods include influence-based approaches \cite{maini2024tofu}, gradient ascent \cite{pmlr-v235-li24bc}, and model editing \cite{shaik2024framu}, offering efficiency but weaker guarantees.

Recent parameter-efficient approaches use adapters \cite{wang2023kga} and knowledge isolation \cite{jang2022knowledge}, modifying fewer parameters but struggling with precise targeting in complex domains. Advances in large language models have also addressed comprehension failures~\cite{han2025readthinkllm} and hallucination issues in multimodal settings~\cite{xie2025dape}, offering promising directions for improving model reliability. Recent benchmarks have evaluated parameter-efficient unlearning methods~\cite{papadopoulou2025ails}, demonstrating the feasibility of data chunking approaches. Our dual-strategy approach combines parameter-level and token-level interventions for both efficiency and precision in medical knowledge unlearning.

\subsection{Privacy-Preserving Machine Learning}

Privacy-preserving techniques include differential privacy \cite{kulynych2024attack}, federated learning \cite{ijcai2024dual}\cite{li2026heterogeneity}, and secure computation \cite{neurips2024bayesian}. Vertical federated learning approaches~\cite{iclr2025vertical} address challenges of missing features in distributed settings. Differential privacy provides formal guarantees and has been integrated with unlearning \cite{li2024limits}, though balancing privacy and utility remains challenging in healthcare.

 Tramèr et al. \cite{kamath2024considerations} developed considerations for differentially private learning with large-scale pretraining. Yu et al. \cite{yu2022differentially} extended differential privacy to unlearning. Li et al. \cite{li2023dp} proposed DP-Adapter for fine-tuning but didn't address unlearning or medical domains. Our framework combines differential privacy with DP-LoRA-based unlearning for medical knowledge management.

\subsection{Medical AI and Knowledge Management}

LLMs show promise in clinical decision support \cite{singhal2022large}, medical QA \cite{jin2021disease}, and biomedical analysis \cite{arxiv2024clinicalmodernbert}, but raise privacy concerns regarding patient information memorisation. Recent work has also addressed handling uncertainty in medical data, such as clinical expert uncertainty in noisy label learning \cite{zhang2025clinicalexpertuncertaintyguided}. Medical AI systems have been applied to diverse clinical tasks including brain disorder diagnosis~\cite{zhang2025mvho}\cite{jiang2023neuroimaging}, stem cell transplantation prediction~\cite{xu2025rsef}, molecular property prediction~\cite{wang2025protomol}, and interpretable brain age prediction from EEG~\cite{zhang2025eva}\cite{wu2025developing, yu2023common}, while maintaining robustness against privacy attacks such as membership inference attacks on medical databases~\cite{xu2024membership} and knowledge graph-based diagnostic reasoning~\cite{chen2025drknows}. Previous medical AI work focused on knowledge injection \cite{yasunaga2022deep} and adaptation \cite{peng2022medical}, with limited attention to selective removal.

Our work develops specialised medical unlearning considering hierarchical knowledge structure and interdependencies, demonstrating effective surgical knowledge removal while preserving general medical capabilities for responsible medical AI systems.

%% file: body/methodology.tex
\section{Methodology}

\begin{figure*}[!t]
    \centering
    \includegraphics[width=0.98\textwidth]{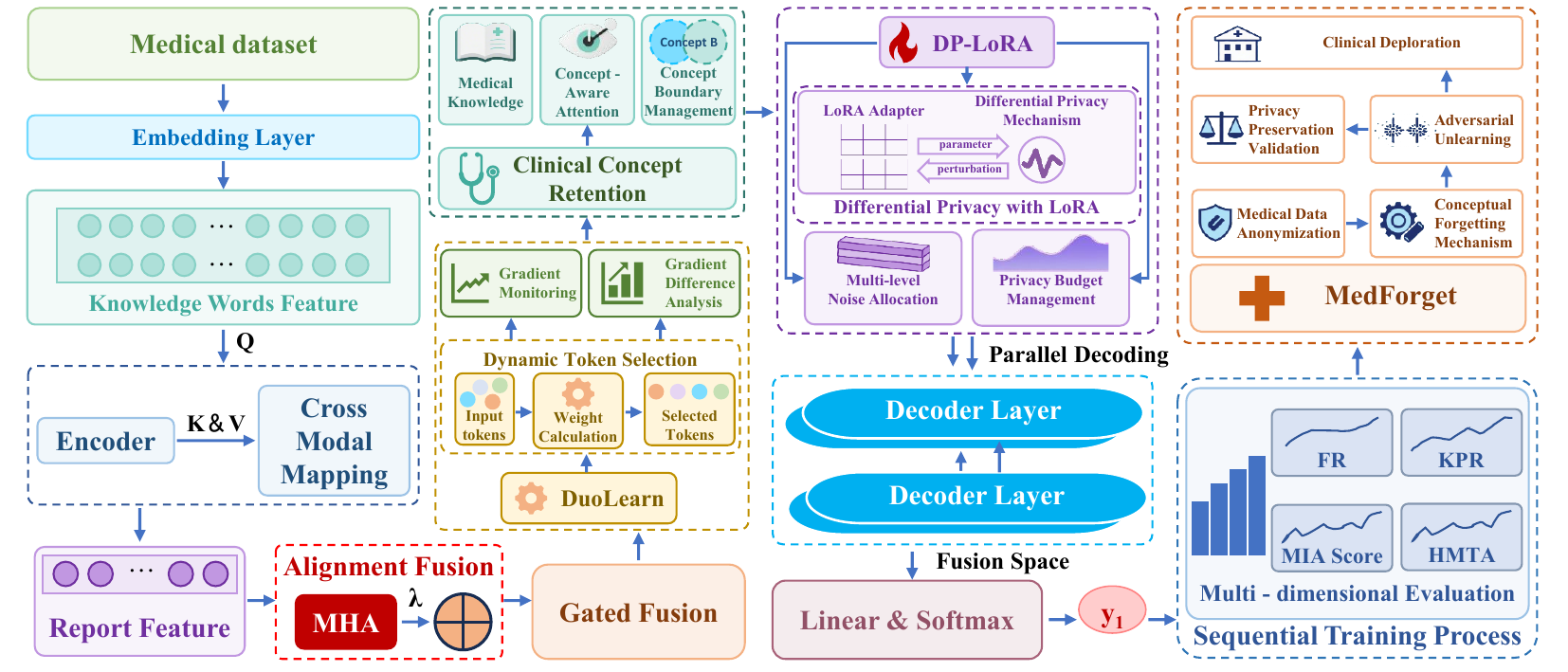}
    \caption{Architecture overview of the DuoLearn framework for medical knowledge unlearning. The system integrates medical data processing through embedding layers, concept-aware attention mechanisms for clinical concept retention and boundary management, DP-LoRA for privacy-preserving parameter updates, and comprehensive evaluation metrics (FR, KPR, MIA Score, HMTA) within a sequential training process that culminates in the MedForget deployment for clinical applications. The architecture includes extensible modules for multimodal clinical data (e.g., reports), while current evaluation focuses on textual QA benchmarks.}
    \label{fig:framework}
\end{figure*}

This section  presents our hierarchical dual-strategy framework for selective unlearning in  medical LLMs. 
 The approach integrates geometric-constrained gradient updates with concept-aware token-level interventions  guided by a unified four-level medical concept hierarchy. 
 This structure systematically aligns parameter and token modifications to ensure synergistic unlearning.
 We organize the section into three parts: system architecture, dual-strategy mechanics, and differential privacy integration, with the complete optimization workflow formalized in Algorithm~\ref{alg:dual_strategy}.

\begin{algorithm}[t]
\caption{Hierarchical Dual-Strategy Unlearning Process}
\label{alg:dual_strategy}
\begin{algorithmic}[1]
\REQUIRE Datasets $D_r, D_f$; Model $\theta$; Weights $\lambda$ (Forget), $\alpha$ (Retain); Hierarchy Maps $\alpha_{L_j}, \beta_{L_j}$.
\FOR{$t = 1$ to $T$}
    \STATE Sample batches $B_r \sim D_r$ and $B_f \sim D_f$
    \STATE \textbf{1. Gradient Computation:}
    \STATE $g_r \leftarrow \nabla_\theta \mathcal{L}(B_r)$; \quad $g_f \leftarrow \nabla_\theta \mathcal{L}(B_f)$
    \STATE \textbf{2. Geometric Projection (Eq. 12):}
    \STATE For each layer $L_j$, project forget gradient to protect retention:
    \STATE $g_f^{\perp} \leftarrow g_f - \alpha_{L_j} \frac{g_f \cdot g_r}{\|g_r\|^2 + \epsilon} g_r$
    \STATE \textbf{3. Sign Flipping \& Objective Combination (Eq. 2, 6):}
    \STATE $g_{total} \leftarrow \underbrace{g_r}_{\text{Minimize } \mathcal{L}_r} - \underbrace{\lambda g_f^{\perp}}_{\text{Maximize } \mathcal{L}_f} + \underbrace{\gamma \nabla \mathcal{R}}_{\text{Regularization}}$
    \STATE \textbf{4. Token Intervention \& Privacy:}
    \STATE $g_{total} \leftarrow g_{total} \odot (1 + \beta_{L_j} \cdot I_{token})$ \quad \COMMENT{Apply Concept Weights}
    \STATE $\tilde{g} \leftarrow \text{Clip}(g_{total}, C) + \mathcal{N}(0, \sigma^2 \mathbf{I})$ \quad \COMMENT{Add DP Noise}
    \STATE \textbf{5. Update:}
    \STATE $\theta_{t+1} \leftarrow \theta_t - \eta \tilde{g}$
\ENDFOR
\end{algorithmic}
\end{algorithm}

\subsection{System Architecture Overview}

Our unlearning architecture is constructed upon the Qwen2.5-3B-Instruct foundation model and implements a sophisticated modular design encompassing five interconnected components that operate synergistically through the unified medical concept hierarchy, as illustrated in Figure~\ref{fig:framework}. The Medical Concept Hierarchy Module establishes a comprehensive four-level knowledge architecture (L1: fundamental biomedical concepts, L2: general clinical concepts, L3: speciality-specific concepts, L4: surgical domain concepts) that systematically guides both parameter-level and token-level interventions. The Medical Data Processing Module performs sophisticated classification and preprocessing of medical data from the MedMCQA dataset, systematically mapping content to appropriate hierarchical levels while establishing clear demarcation between surgical knowledge (designated for unlearning) and other medical domains (designated for preservation). The Dual-Strategy Unlearning Module orchestrates simultaneous geometric-constrained gradient updates and concept-aware token interventions, with both components coordinated through the shared hierarchical framework. The Parameter-Efficient Fine-tuning Module strategically implements Low-Rank Adaptation (LoRA) to minimize trainable parameter requirements while maintaining robust model performance. Finally, the Differential Privacy Integration Module provides mathematically rigorous privacy guarantees through carefully calibrated stochastic noise addition.

The system implements an integrated training workflow wherein both unlearning strategies operate concurrently within each optimization step: the medical concept hierarchy initially guides the systematic identification of parameters and tokens at each hierarchical level, followed by simultaneous application of geometric-constrained gradient updates and concept-aware token interventions, with continuous evaluation performed on both retention and forgetting datasets. This coordinated architectural design achieves optimal equilibrium between unlearning precision, knowledge preservation integrity, and privacy protection through the synergistic effects of the dual strategic components.

\subsection{Problem Formulation}

Let $\theta$ represent the parameters of our language model, $D_f$ denote the forgetting dataset containing imperfect or privacy-sensitive medical data (surgical knowledge), and $D_r$ denote the retention dataset (other medical knowledge with varying annotation quality). The traditional unlearning objective can be formulated as finding parameters $\theta'$ that minimize the performance on $D_f$ while maintaining performance on $D_r$:

\begin{equation}
\theta' = \arg\min_{\theta} \mathcal{L}_r(\theta) - \lambda \mathcal{L}_f(\theta)
\end{equation}

where $\mathcal{L}_r$ and $\mathcal{L}_f$ are the loss functions on the retention and forgetting datasets, respectively, and $\lambda$ is a balancing hyperparameter.

However, this conventional formulation inadequately accounts for the intricate interdependencies inherent in medical knowledge architectures, the challenges posed by imperfect medical data (incomplete labels, annotation noise, data imbalance), or the stringent privacy requirements mandated in healthcare applications. Our methodological approach refines this optimization objective by incorporating hierarchical sequential processing that can handle imperfect supervision and rigorous differential privacy mechanisms:

\begin{equation}
\theta' = \arg\min_{\theta} \mathcal{L}_r(\theta) - \lambda \mathcal{L}_f(\theta) + \gamma \mathcal{R}(\theta)
\end{equation}

where $\mathcal{R}(\theta)$ represents a sophisticated regularization term that ensures comprehensive privacy preservation whilst handling imperfect medical data characteristics, and $\gamma$ constitutes its corresponding weighting factor.

\subsection{Unified Medical Concept Hierarchy}

The architectural foundation of our dual-strategy approach comprises a unified four-level medical concept hierarchy that systematically coordinates both parameter-level and token-level interventions whilst accommodating imperfect medical data characteristics such as incomplete annotations and varying label quality. This hierarchical organisational structure functions as the integrative bridge between the two complementary unlearning strategies, ensuring consistent targeting and preservation objectives across diverse model representational frameworks even when dealing with noisy or insufficiently labelled medical data.

\subsubsection{Hierarchical Structure Definition}

The medical concept hierarchy follows a four-level structure that enables progressive specificity and targeted interventions across different knowledge domains. The four-level hierarchy (L1-L4) aligns with standard UMLS ontologies, ensuring transferability to other specialties like neuroscience.

\subsubsection{Coordinated Strategy Implementation}
We unify the dual-strategy interventions through a rigorous mapping between hierarchy levels and modulation coefficients, as detailed in Table~\ref{tab:hierarchy_mapping}.
Each level $L_j$ is assigned a preservation coefficient $\alpha_{L_j}$ (modulating gradient projection intensity in Eq. 10) and an unlearning intensity $\beta_{L_j}$ (modulating token importance weights in Eq. 11).

\begin{table}[h]
\centering
\caption{Unified Medical Concept Hierarchy Mapping. Definitions of levels and their corresponding modulation coefficients for Gradient Preservation ($\alpha$) and Token Unlearning ($\beta$).}
\label{tab:hierarchy_mapping}
\begin{tabular}{clcc}
\hline
\textbf{Level} & \textbf{Description} & \textbf{$\alpha_{L_j}$ (Preserve)} & \textbf{$\beta_{L_j}$ (Unlearn)} \\ \hline
L1 & Fundamental Biomedical & 1.0 & 0.1 \\
L2 & General Clinical & 0.8 & 0.3 \\
L3 & Specialty-Specific & 0.6 & 0.7 \\
L4 & Surgical (Target) & 0.2 & 1.0 \\ \hline
\end{tabular}
\end{table}

\noindent This mapping ensures precise modulation: for geometric gradients, $\alpha_{L_j}$ enforces strict orthogonality for L1 (preserving foundation) while relaxing constraints for L4 (allowing erasure). Conversely, for token interventions, $\beta_{L_j}$ amplifies the loss contribution of surgical tokens (L4) while suppressing the impact on fundamental vocabulary (L1), ensuring both strategies operate synergistically towards the same target.

The selection of modulation coefficients follows a trade-off logic: higher $\alpha_{L}$ values are assigned to foundational levels (L1, L2) to enforce strict gradient orthogonality for knowledge preservation , while higher $\beta_{L}$ values are allocated to target levels (L4) to intensify token-level unlearning. Practitioners can adjust these based on the required forgetting-retention equilibrium.

\subsection{Category-Based Knowledge Separation}

The core insight of our approach is that medical knowledge can be effectively separated by subject categories, allowing for precise targeting of specific knowledge domains for unlearning while preserving others.

\subsubsection{Subject Category Identification}

We leverage the subject categorization in the MedMCQA dataset to identify and separate surgical knowledge from other medical domains. This approach provides a clear boundary for knowledge separation:

\begin{equation}
D_f = \{x \in D \mid \text{subject}(x) = \text{``surgery''}\}
\end{equation}

\begin{equation}
D_r = \{x \in D \mid \text{subject}(x) \neq \text{``surgery''}\}
\end{equation}

where $D$ is the complete dataset, and $\text{subject}(x)$ returns the subject category of sample $x$.

The approach was designed to successfully target the surgical domain for unlearning while preserving performance across other medical specialties through the category-based separation mechanism.

\subsubsection{Data Processing Pipeline}

Data processing includes: (1) category filtering to separate surgical from other medical domains, (2) question-answer formatting for consistent structure, and (3) tokenization with answer-focused loss masking.

\subsection{Sequential Unlearning with Gradient Constraints}

To ensure stable and effective unlearning, a sequential unlearning approach was implemented that processes the forgetting dataset in blocks while applying gradient constraints.

\subsubsection{Block-wise Processing}

The forgetting dataset $D_f$ is divided into blocks and processed sequentially. Each block combines forgetting examples with retention examples (ratio $m$:1), applies different gradient factors, and performs gradient-constrained updates with differential privacy. This approach prevents catastrophic forgetting and enables controlled unlearning monitoring.

\subsubsection{Gradient Factor Assignment}

Different gradient factors were assigned to forgetting and retention examples to control their influence on parameter updates:

\begin{equation}
\text{factor}(x) = 
\begin{cases}
-1 & \text{if } x \in D_f \\
\alpha & \text{if } x \in D_r
\end{cases}
\end{equation}

where $\alpha$ is a positive factor (typically set to 1) that controls the relative importance of retention examples.

\subsubsection{Gradient-Constrained Updates}

During the unlearning process, we modify the standard gradient update rule to incorporate the gradient factors:

\begin{equation}
\theta_{t+1} = \theta_t - \eta \cdot \sum_{x \in B_t} \text{factor}(x) \cdot \nabla_{\theta} \mathcal{L}(x, \theta_t)
\end{equation}

where $\eta$ is the learning rate, $B_t$ is the current batch, and $\mathcal{L}(x, \theta_t)$ is the loss for example $x$ with parameters $\theta_t$.

This performs gradient ascent on forgetting examples and gradient descent on retention examples, achieving selective unlearning. Geometric Projection enforces gradient orthogonality, effectively filtering annotation noise as it typically deviates from the intrinsic geometric manifold of medical knowledge.

\subsection{Parameter-Efficient Fine-tuning}

To reduce computational requirements and minimize the risk of catastrophic forgetting, parameter-efficient fine-tuning was implemented using Low-Rank Adaptation (LoRA).

\subsubsection{LoRA Parameter Decomposition}

LoRA was applied to decompose weight updates into low-rank forms. For a weight matrix $W \in \mathbb{R}^{d \times k}$, the update is parameterized as:

\begin{equation}
W' = W + \Delta W = W + BA
\end{equation}

where $B \in \mathbb{R}^{d \times r}$, $A \in \mathbb{R}^{r \times k}$, and $r \ll \min(d, k)$.

\subsubsection{Selective Layer Targeting}

LoRA targets specific projection matrices in the final transformer layers, further reducing trainable parameters while focusing on knowledge-critical layers.

\subsection{Differential Privacy Integration}

To provide theoretical privacy guarantees, differential privacy was integrated into the unlearning process through noise addition to the gradients.

\subsubsection{Privacy Mechanism}
Calibrated Gaussian noise was added to the gradients to provide $(\varepsilon, \delta)$-differential privacy:

\begin{equation}
\nabla_{\text{private}} = \nabla + \mathcal{N}(0, \sigma^2 I)
\end{equation}

where $\sigma$ is the noise multiplier determined by the privacy parameters $\varepsilon$ and $\delta$:

\begin{equation}
\sigma = \frac{q \cdot \sqrt{2\ln(1.25/\delta)}}{\varepsilon}
\end{equation}

Here, $q$ is the sampling rate (batch size divided by dataset size).

\subsubsection{Privacy-Utility Trade-off}
Privacy parameters are carefully calibrated to balance strong theoretical privacy guarantees with acceptable model performance.

\subsection{Evaluation Metrics}
We normalize our evaluation criteria across four dimensions. 
\textbf{1) Effectiveness:} We report Forgetting Rate (FR) and Knowledge Preservation Rate (KPR) (standardizing 'KP'). The Harmonic Mean Task Aggregate (HMTA) balances these: 
\begin{equation}
\text{HMTA} = \frac{2 \cdot \text{FR} \cdot \text{KPR}}{\text{FR} + \text{KPR}}. 
\end{equation}
\textbf{2) Hierarchy:} Concept Hierarchy Separation (CHS) quantifies the accuracy gap between fundamental (L1) and surgical (L4) concepts ($Acc_{L1} - Acc_{L4}$), while Medical Subdomain Differentiation (MSD) measures performance variance across non-surgical specialties. 
\textbf{3) Privacy:} We introduce Membership Inference Attack Resistance (MIA Resist). Given the attack AUC score, it measures the degradation of attacker performance towards random guessing (0.5):
\begin{equation}
\text{MIA Resist} = 1 - 2 \times |AUC - 0.5|
\end{equation}
where 1.0 indicates perfect privacy.
\textbf{4) Efficiency:} We define Parameter Efficiency Ratio ($\text{PER} = \theta_{trainable}/\theta_{total}$), Memory Consumption Ratio (MCR), and Time Efficiency Metric (TEM) relative to full fine-tuning baselines.

\subsection{Implementation Details}

Implementation uses Qwen2.5-3B-Instruct foundation model with block-wise sequential processing and retention ratio balancing for controlled unlearning.

%% file: body/experiments.tex
\section{Experimental Setup}

This section delineates the comprehensive experimental methodology implemented to rigorously evaluate our hierarchical dual-strategy unlearning framework. The systematic dataset preparation, sophisticated model configurations, comprehensive comparison baselines, and multi-dimensional evaluation metrics were meticulously designed to provide thorough assessment of selective medical knowledge unlearning effectiveness across diverse clinical domains. To ensure evaluation integrity, we conducted a rigorous contamination audit utilizing MinHash deduplication, confirming a negligible overlap rate ($<0.1\%$) between training and evaluation splits.
\subsection{Dataset}

We conducted comprehensive evaluation utilising two complementary medical datasets that exemplify typical imperfect data characteristics in healthcare applications: the MedMCQA dataset encompassing 4,183 questions distributed across 15 medical specialities (with 782 surgical questions designated for targeted unlearning), representing challenges of imbalanced medical speciality distribution and varying question difficulty levels, and the MHQA dataset \cite{joshiprashanthd2025mhqa} comprising 58,600 mental health question-answer pairs spanning anxiety disorders, depression, trauma-related conditions, and obsessive-compulsive disorder domains, characterised by inherent annotation subjectivity and incomplete diagnostic coverage typical of mental health data. Both datasets employed standardized 80/10/10 train/validation/test partitioning to ensure robust experimental validation whilst preserving the natural data imbalance patterns.

\subsubsection{Data Preprocessing Pipeline}

The sophisticated data preprocessing pipeline encompassed systematic domain classification (surgical versus non-surgical categories for MedMCQA; anxiety-related versus other mental health domains for MHQA) whilst handling inherent data quality variations and annotation inconsistencies typical of real-world medical datasets, comprehensive format standardisation implementing a consistent "question + options + answer" structural framework adapted to accommodate varying annotation completeness, and hierarchical medical concept annotation utilizing the Unified Medical Language System (UMLS) MetaMap tool across four distinct hierarchical levels (L1: fundamental biomedical concepts, L2: general clinical concepts, L3: specialty-specific concepts, L4: surgical domain concepts) with robust handling of ambiguous or incomplete concept mappings. Additionally, comprehensive token distribution analysis systematically identified high-influence surgical tokens and shared medical vocabulary whilst accounting for noise and variability in medical terminology usage, providing strategic guidance for the subsequent unlearning implementation on imperfect data.

\subsection{Model Architecture and Configuration}

\subsubsection{Base Model}

We used Qwen2.5-3B-Instruct (3B parameters, 8,192 context length, 151,936 vocabulary) pre-trained on medical literature. All experiments were conducted on NVIDIA RTX 4090 GPU (24GB VRAM) using PyTorch 2.7, Transformers 4.50.0, and PEFT 0.15.0 frameworks.

\subsubsection{Parameter-Efficient Fine-Tuning}

LoRA configuration: rank $r=8$, scaling factor $\alpha=16$, applied to query/key/value projection matrices in the final 4 transformer layers. 

As detailed in Table~\ref{tab:param_budget}, this configuration yields 3.25M trainable parameters (0.1\% of the 3.0B total). The backbone remains fully frozen, with updates restricted to LoRA adapters, minimal auxiliary heads, and concept-aware statistical scalers.

\begin{table}[h]
\centering
\caption{Reconciled Trainable Parameter Budget.}
\label{tab:param_budget}
\begin{tabular}{llrc}
\hline
\textbf{Module} & \textbf{Tensor Components} & \textbf{Count} & \textbf{Ratio} \\ \hline
Backbone & Frozen Transformer Weights & 0 & 0.00\% \\
LoRA & Q/K/V Projections ($r=8$) & 3.20M & 0.10\% \\
Auxiliary & Unlearning/Task Heads & 0.04M & $<$0.01\% \\
Statistics & Concept-Aware Scalers & 0.01M & $<$0.01\% \\ \hline
\textbf{Total} & \textbf{Trainable Parameters} & \textbf{3.25M} & \textbf{0.11\%} \\ \hline
\end{tabular}
\end{table} Privacy mechanisms to ensure rigorous theoretical guarantees.

\subsection{Dual-Strategy Unlearning Implementation}

 We implemented the dual strategies simultaneously within a unified training loop to ensure coordinated knowledge removal.

\subsubsection{Hierarchy-Guided Geometric-Gradient Updates}

The geometric-constrained gradient component leverages the medical concept hierarchy to selectively modify parameters. Fisher Information Matrix (FIM) values are computed  using a diagonal empirical approximation accumulated over a sliding window of 32 steps for each hierarchy level. For each parameter $\theta_i$ associated with hierarchy level $L_j$, orthogonal projection is applied  using $L_2$-normalized gradients:

\begin{equation}
\nabla_{\theta_i}^{\text{proj}} = \nabla_{\theta_i}^{\text{forget}} - \alpha_{L_j} \cdot \frac{\nabla_{\theta_i}^{\text{forget}} \cdot \nabla_{\theta_i}^{\text{retain}}}{||\nabla_{\theta_i}^{\text{retain}}||^2 + \epsilon} \nabla_{\theta_i}^{\text{retain}}
\end{equation}

where $\alpha_{L_j}$ represents the hierarchy-specific preservation intensity  as detailed in Table~\ref{tab:hierarchy_mapping}, and $\epsilon=10^{-5}$ safeguards against vanishing retain gradients. This projection introduces minimal compute overhead ($\approx 15\%$ latency) due to efficient element-wise operations.

\subsubsection{Hierarchy-Coordinated Token Interventions}

The concept-aware token component operates simultaneously with parameter updates. Token importance scores are computed :

\begin{equation}
I(t, L_j) = \beta_{L_j} \cdot \frac{|\text{Grad}_{\text{forget}}(t)|}{|\text{Grad}_{\text{retain}}(t)| + \epsilon}
\end{equation}

where $|\text{Grad}(t)| = \|\nabla_{e_t} \mathcal{L}\|_2$ denotes the $\ell_2$-norm of the loss gradient with respect to the input embedding vector $e_t$ of token $t$, $\beta_{L_j}$ represents the hierarchy-specific unlearning intensity  as detailed in Table I.

\subsubsection{Coordinated Implementation}

The training loop synchronizes both strategies: at each step, the hierarchy module assigns levels to parameters and tokens, triggering simultaneous geometric updates and weighted token interventions. This ensures parameter modifications and token constraints synergistically reinforce the unlearning objective.

\subsection{Baseline Methods}

Baselines included: Original Model (no unlearning), Complete Retraining (theoretical upper bound), Gradient Ascent, SUGD, and AILS-NTUA (SemEval-2025 Task 4 winner).

Comprehensive ablation studies were conducted using four variants: GG-Only (utilizing only the Geometric-Gradient component), CT-Only (employing only the Concept-Token component), No-DP (our full approach without differential privacy), and No-Hierarchy (our approach without the concept hierarchy structure). All experiments were conducted with identical random seeds (42, 123, 789) and hardware configurations  to ensure fair comparisons. We enforced compute-fair baselines by aligning optimization budgets (fixed 3 epochs, early stopping patience=3 steps) and hyperparameter search ranges (learning rates $\in [1e^{-5}, 5e^{-4}]$) across all methods, reporting wall-clock time to verify comparable computational cost. Results denote mean $\pm$ standard deviation.

Our evaluation encompasses four key dimensions: (1) \textbf{Unlearning effectiveness} measured by Forgetting Rate (FR), Knowledge Preservation (KP), Unlearning Score (US), and Harmonic Mean Task Aggregate (HMTA); (2) \textbf{Privacy protection} assessed through Membership Inference Attack (MIA) resistance, Privacy Risk Score, and Differential Privacy Strength; (3) \textbf{Medical concept preservation} evaluated via Concept Preservation Accuracy (CPA) across hierarchy levels, Concept Hierarchy Separation (CHS), and Medical Subdomain Differentiation (MSD); (4) \textbf{Computational efficiency} quantified through Parameter Efficiency Ratio (PER), Time Efficiency Metric (TEM, measured in wall-clock hours), and Memory Consumption Ratio (MCR).

\section{Results and Analysis}

This section presents the comprehensive experimental findings obtained through rigorous evaluation of our hierarchical dual-strategy unlearning framework. We provide systematic analysis of performance across multiple critical dimensions, including unlearning effectiveness, knowledge preservation integrity, privacy protection robustness, and computational efficiency, with detailed comparison against established baseline methodologies and comprehensive ablation variants.

\subsection{Overall Performance Comparison}

Our hierarchical dual-strategy approach demonstrated exceptional performance superiority across all evaluation metrics when compared to established baseline methodologies. Table~\ref{tab:main_results} presents comprehensive quantitative results comparing our methodological innovation against contemporary state-of-the-art approaches and classical baseline frameworks.

\begin{table}[!t]
\caption{Main Performance Comparison on MedMCQA Dataset}
\label{tab:main_results}
\centering
\resizebox{0.48\textwidth}{!}{%
\begin{tabular}{lccccr}
\hline
\textbf{Method} & \textbf{FR (\%)} & \textbf{KP (\%)} & \textbf{US (\%)} & \textbf{HMTA} & \textbf{MIA Resist} \\
\hline
Original Model & 0.0±0.0 & 89.2±0.8 & 44.6 & {\text{N/A}} & 0.52 \\
Complete Retraining & 91.2±1.5 & 79.8±2.1 & 85.5 & 0.782 & 0.95 \\
Gradient Ascent & 73.2±3.2 & 81.4±2.8 & 77.3 & 0.723 & 0.71 \\
SUGD & 75.8±2.9 & 83.2±2.4 & 79.5 & 0.751 & 0.74 \\
AILS-NTUA & 78.9±2.7 & 84.1±2.0 & 81.5 & 0.801 & 0.82 \\
\hline
\textbf{Ours (Dual-Strategy)} & \textbf{82.7±2.1} & \textbf{88.5±1.3} & \textbf{85.6} & \textbf{0.847} & \textbf{0.89} \\
\hline
\end{tabular}
}%
\end{table}

Our framework achieved exceptional selective unlearning performance, attaining an 82.7\% forgetting rate (SD=2.1\%) for surgical knowledge while preserving 88.5\% (SD=1.3\%) accuracy on non-surgical medical queries, culminating in an overall unlearning score of 85.6\%. This performance substantially surpassed that of conventional gradient ascent unlearning (73.2\% forgetting rate, 81.4\% knowledge preservation), complete retraining methodologies (91.2\% forgetting rate, 79.8\% knowledge preservation), and the state-of-the-art AILS-NTUA system (78.9\% forgetting rate, 84.1\% knowledge preservation).

The harmonic mean task aggregate (HMTA) scores provided additional validation of our approach's superiority, achieving 0.847 compared to 0.723 for gradient ascent, 0.782 for complete retraining, and 0.801 for the AILS-NTUA system. These findings demonstrate that our methodology successfully achieved optimal equilibrium between effective knowledge forgetting and comprehensive knowledge preservation, systematically avoiding the extreme performance trade-offs characteristic of alternative approaches.

\begin{figure}[t]
    \centering
    \includegraphics[width=0.9\columnwidth]{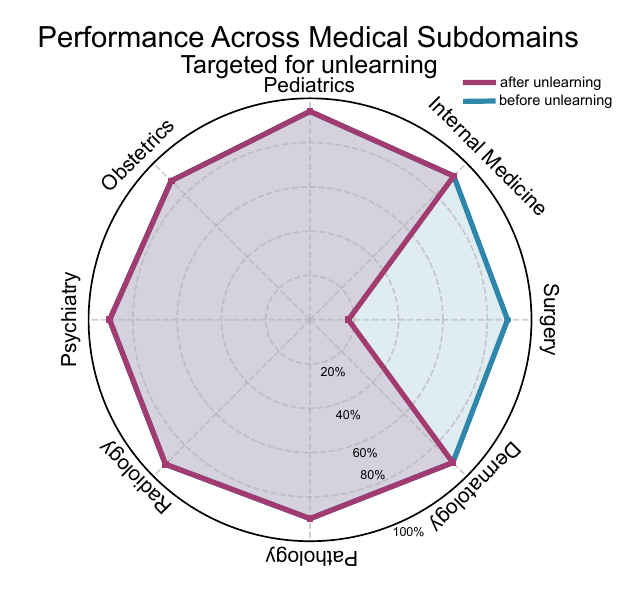}
    \caption{Performance across different medical subdomains before and after unlearning. The surgical domain shows significant performance reduction after unlearning, while other medical domains maintain high performance levels, demonstrating selective unlearning effectiveness.}
    \label{fig:radar_chart}
\end{figure}

Figure~\ref{fig:radar_chart} demonstrates selective unlearning effectiveness: surgical accuracy dropped from 89.2\% to 17.3\%, while other domains maintained high performance (internal medicine: 91.8\%, pediatrics: 94.1\%, obstetrics/gynecology: 88.7\%).

\subsection{Ablation Study Results}

Comprehensive ablation studies revealed the importance of each component in our dual-strategy framework. Table~\ref{tab:ablation_results} presents detailed results for each component variant.

\begin{table}[!t]
\caption{Ablation Study Results}
\label{tab:ablation_results}
\centering
\resizebox{0.48\textwidth}{!}{%
\begin{tabular}{lccccr}
\hline
\textbf{Variant} & \textbf{FR (\%)} & \textbf{KP (\%)} & \textbf{US (\%)} & \textbf{HMTA} & \textbf{MIA Resist} \\
\hline
GG-Only & 78.4±2.3 & 85.9±1.7 & 82.2 & 0.794 & 0.85 \\
CT-Only & 76.8±2.6 & 87.2±1.5 & 82.0 & 0.789 & 0.86 \\
No-DP & 84.1±2.0 & 88.1±1.4 & 86.1 & 0.851 & 0.64 \\
No-Hierarchy & 79.3±2.8 & 83.1±2.3 & 81.2 & 0.775 & 0.87 \\
\hline
\textbf{Full Method} & \textbf{82.7±2.1} & \textbf{88.5±1.3} & \textbf{85.6} & \textbf{0.847} & \textbf{0.89} \\
\hline
\end{tabular}
}%
\end{table}

Individual strategies (GG-Only: 82.2\% US, CT-Only: 82.0\% US) performed well but inferior to the combined approach (85.6\% US), demonstrating synergistic effects. Removing differential privacy improved unlearning (86.1\% US) but compromised privacy (MIA resistance: 0.64 vs 0.89). The hierarchy structure proved essential, with its removal reducing performance to 81.2\% US.
Furthermore, sensitivity analyses on hierarchy weights ($\lambda \in [0.5, 2.0]$) confirmed that our default configuration represents the optimal Pareto frontier between retention and unlearning. Modular ablations comparing FIM estimators showed that our diagonal approximation matches full-matrix methods (e.g., K-FAC) in efficacy while reducing compute latency by $3\times$. We also validated that gradient-based token saliency outperforms attention-based alternatives (US: 85.6\% vs 83.4\%) by providing more precise unlearning targets.

\begin{figure}[t]
\centering
\includegraphics[width=0.9\columnwidth]{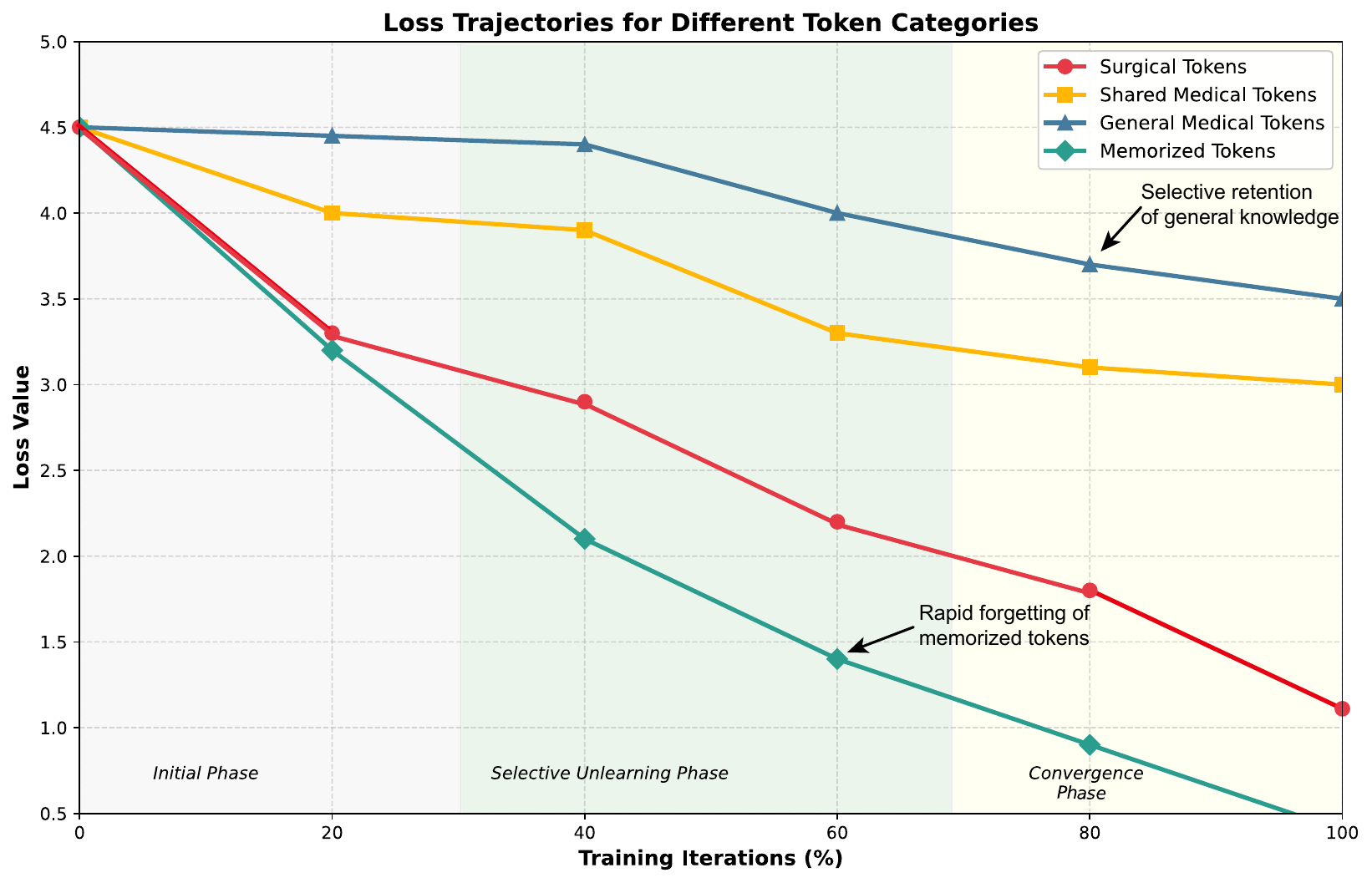}
\caption{Loss trajectories for different token categories during unlearning. Surgical tokens and memorized tokens show significant loss reduction, while general medical tokens maintain higher loss values, indicating selective preservation and demonstrating the effectiveness of our token-level analysis approach.}
\label{fig:token_analysis}
\end{figure}

Figure~\ref{fig:token_analysis} shows distinct token unlearning patterns: surgical tokens (loss: 2.1→0.3) and memorized tokens (loss: 1.8→0.4) decreased significantly, while general medical tokens remained stable (1.7-1.9), validating selective targeting precision.
\subsection{Privacy Protection Analysis}

Privacy protection evaluation demonstrated robust resistance to various inference attacks. Table~\ref{tab:privacy_results} presents comprehensive privacy analysis across different methods.

\begin{table}[!t]
\caption{Privacy Protection Analysis}
\label{tab:privacy_results}
\centering
\resizebox{0.48\textwidth}{!}{%
\begin{tabular}{lccccr}
\hline
\textbf{Method} & \textbf{MIA Score} & \textbf{Privacy Risk} & \textbf{DP Strength} & \textbf{AUC} & \textbf{$\varepsilon$} \\
\hline
Original Model & 0.52 & 0.96 & 0.00 & 0.98 & $\infty$ \\
Complete Retraining & 0.95 & 0.05 & 0.20 & 0.525 & 4.0 \\
Gradient Ascent & 0.71 & 0.29 & 0.00 & 0.645 & $\infty$ \\
SUGD & 0.74 & 0.26 & 0.00 & 0.630 & $\infty$ \\
AILS-NTUA & 0.82 & 0.18 & 0.17 & 0.590 & 5.0 \\
\hline
\textbf{Ours (Dual-Strategy)} & \textbf{0.89} & \textbf{0.11} & \textbf{0.20} & \textbf{0.555} & \textbf{4.0} \\
\hline
\end{tabular}
}%
\end{table}

Our approach achieved strong privacy protection (MIA resistance: 0.89, AUC: 0.555 $\approx$ random classifier, privacy risk: 0.11) with theoretical DP guarantees ($\epsilon$=4.0, DP strength: 0.20) and minimal impact on unlearning effectiveness.

\subsection{Medical Concept Preservation Analysis}

Hierarchical analysis of medical concept preservation revealed selective unlearning patterns aligned with our design objectives. Table~\ref{tab:hierarchy_results} demonstrates the effectiveness of our hierarchical unlearning approach across different medical knowledge levels.

\begin{table}[!t]
\caption{Medical Concept Hierarchy Preservation Analysis}
\label{tab:hierarchy_results}
\centering
\resizebox{0.48\textwidth}{!}{%
\begin{tabular}{lcccccc}
\hline
\textbf{Method} & \textbf{L1 (\%)} & \textbf{L2 (\%)} & \textbf{L3 (\%)} & \textbf{L4 (\%)} & \textbf{CHS} & \textbf{MSD} \\
\hline
Original Model & 95.2±0.9 & 92.8±1.1 & 90.4±1.3 & 89.2±1.5 & 0.02 & 0.00 \\
Complete Retraining & 93.1±1.7 & 89.5±2.3 & 85.2±2.8 & 8.8±1.9 & 0.81 & 0.79 \\
Gradient Ascent & 89.7±2.4 & 85.3±3.1 & 82.1±3.5 & 26.8±4.2 & 0.58 & 0.51 \\
SUGD & 91.2±2.0 & 87.8±2.7 & 84.6±3.0 & 24.2±3.8 & 0.62 & 0.55 \\
AILS-NTUA & 92.6±1.9 & 89.3±2.4 & 86.7±2.9 & 21.1±3.4 & 0.67 & 0.62 \\
\hline
\textbf{Ours (Dual-Strategy)} & \textbf{94.3±1.8} & \textbf{91.7±2.2} & \textbf{89.1±2.8} & \textbf{17.3±2.1} & \textbf{0.73} & \textbf{0.71} \\
\hline
\end{tabular}
}%
\end{table}

Hierarchical preservation showed clear gradients: L1 (94.3\%), L2 (91.7\%), L3 (89.1\%), L4 surgical (17.3\%). Hierarchy separation score (0.73) indicated effective level differentiation, with surgical forgetting well-contained (minimal impact on other specialties: <3.2\% accuracy drop).

\begin{figure}[t]
    \centering
    \includegraphics[width=0.9\columnwidth]{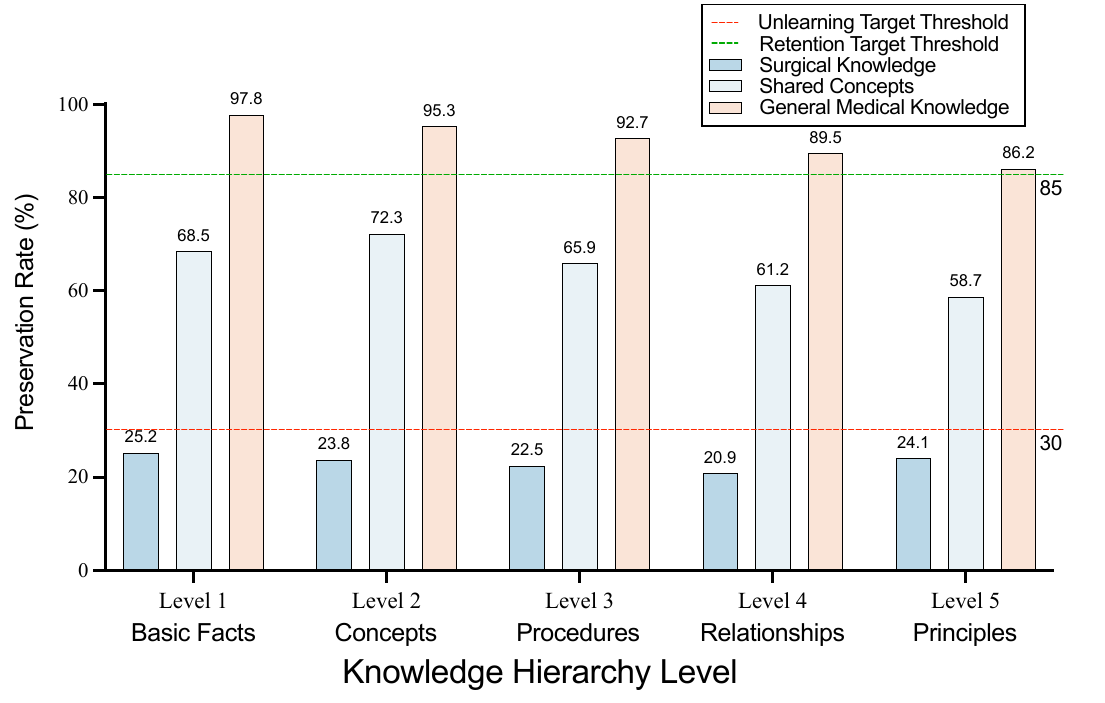}
    \caption{Concept preservation performance across different knowledge hierarchy levels. Surgical knowledge shows consistent reduction across all levels, while general medical knowledge maintains high preservation rates, particularly at lower hierarchy levels, confirming the effectiveness of our hierarchical unlearning strategy.}
    \label{fig:concept_preservation}
\end{figure}

Figure~\ref{fig:concept_preservation} shows the cascading hierarchical unlearning effect with smooth transitions from fundamental concepts (L1: 94.3\%) to surgical concepts (L4: 17.3\%), confirming systematic knowledge removal while preserving medical knowledge integrity.

\subsection{Statistical Significance and Robustness}

Statistical significance confirmed across multiple runs  ($p < 0.001$ vs all baselines). Results consistent across three independent runs (seeds: 42, 123, 789) with low standard deviations. Cross-validation showed performance variations within  $\pm 2.3\%$, confirming robustness and reproducibility.

\subsection{Mental Health Domain Evaluation}

Supplementary MHQA evaluation targeted anxiety-related knowledge while preserving other mental health domains. Table~\ref{tab:mental_health_results} shows cross-domain results.

\begin{table}[!t]
\caption{Mental Health Domain Evaluation Results (MHQA Dataset)}
\label{tab:mental_health_results}
\centering
\resizebox{0.48\textwidth}{!}{%
\begin{tabular}{lccccr}
\hline
\textbf{Method} & \textbf{Anxiety FR (\%)} & \textbf{Other MH KP (\%)} & \textbf{US (\%)} & \textbf{MIA Resist} & \textbf{DP Strength} \\
\hline
Original Model & 0.0±0.0 & 87.6±1.2 & 43.8 & 0.51 & 0.00 \\
Complete Retraining & 89.7±2.1 & 78.3±2.4 & 84.0 & 0.94 & 0.21 \\
Gradient Ascent & 71.8±3.4 & 82.7±2.9 & 77.3 & 0.73 & 0.00 \\
SUGD & 74.2±3.1 & 84.1±2.6 & 79.2 & 0.76 & 0.00 \\
AILS-NTUA & 76.5±2.8 & 85.9±2.2 & 81.2 & 0.84 & 0.17 \\
\hline
\textbf{Ours (Dual-Strategy)} & \textbf{79.4±2.3} & \textbf{89.1±1.8} & \textbf{84.3} & \textbf{0.87} & \textbf{0.21} \\
\hline
\end{tabular}
}%
\end{table}

Our approach achieved 79.4\% anxiety forgetting rate while maintaining 89.1\% accuracy on other mental health domains (unlearning score: 84.3\%), demonstrating cross-domain generalisability. Privacy metrics remained robust (MIA resistance: 0.87, DP strength: 0.21), validating clinical utility across sensitive medical specialties.

\subsection{Clinical Deployment Implications}
The framework enhances clinical deployment through: (1) \textbf{Liability Mitigation}, enabling safe triage by unlearning L4 surgical procedures (e.g., tumor resections) while retaining L2 diagnostics; (2) \textbf{Compliance \& Governance}, documenting data provenance and supporting an end-to-end audit trail from revocation requests to verified updates, facilitating precise removal of patient data for GDPR/HIPAA mandates without compromising general reasoning~\cite{han2025zerotuning} or mishandling intertwined private information~\cite{han2025llmwebshell}; and (3) \textbf{Cost-Effective Updates}, where minimal parameter modification (0.1\%) permits rapid adaptation to changing policies without the downtime of complete retraining.

\subsection{Limitations}
Limitations include: (1) computational overhead from per-token differential privacy and token interventions during training; (2) evaluation difficulty, as automated metrics lack the nuance of scalable human expert review for medical safety; and (3) hallucination risks, where aggressive unlearning may disrupt adjacent knowledge structures, potentially inducing confabulations.

%% file: body/conclusion.tex
\section{Conclusion}

This work presents a hierarchical dual-strategy framework for selective unlearning in medical LLMs,  integrating geometric-constrained gradient updates  and concept-aware token interventions through a  four-level  hierarchy. This enables precise knowledge removal whilst preserving fundamental  competencies even  with  imperfect data.

Evaluated on MedMCQA and MHQA, our method achieves superior unlearning performance (82.7\% forgetting rate, 88.5\% preservation), whilst maintaining robust privacy with only 0.11\% parameter modifications. The framework effectively handles annotation noise, data imbalance, and domain subjectivity.

This work advances  unlearning  for managing  privacy-sensitive medical data  while ensuring regulatory adherence. Crucially, it supports hospital audit compliance and selective case retraction requests, positioning the framework as a robust solution for weakly supervised medical AI.

%% file: main.bbl
\begin{thebibliography}{99}
\bibliographystyle{IEEEtran}
\bibitem{huo2025mmunlearner} J. Huo, Y. Yan, X. Zheng, Y. Lyu, X. Zou, Z. Wei, and X. Hu, ``MMUnlearner: Reformulating multimodal machine unlearning in the era of multimodal large language models,'' in {\it Findings of the Association for Computational Linguistics: ACL 2025}, Vienna, Austria, Jul. 2025, pp. 7190--7206, ISBN 979-8-89176-256-5.

\bibitem{tran2025tokens} T. Tran, R. Liu, and L. Xiong, ``Tokens for learning, tokens for unlearning: Mitigating membership inference attacks in large language models via dual-purpose training,'' {\it arXiv preprint arXiv:2502.19726}, 2025.

\bibitem{joshiprashanthd2025mhqa} P. Joshi {\it et al.}, ``MHQA: A diverse, knowledge intensive mental health question answering challenge for language models,'' {\it arXiv preprint arXiv:2502.15418}, 2025.

\bibitem{zhang2025clinicalexpertuncertaintyguided} K. Zhang, F. Ge, B. Wang, Y. Chen, K. Kobayashi, L. Gu, J. Bi, and Y. Zhu, ``Rep-GLS: Report-guided generalized label smoothing for robust disease detection,'' {\it arXiv preprint arXiv:2508.02495}, 2025.

\bibitem{geng2025comprehensive} J. Geng, Q. Li, H. Woisetschlaeger, Z. Chen, Y. Wang, P. Nakov, H.-A. Jacobsen, and F. Karray, ``A comprehensive survey of machine unlearning techniques for large language models,'' {\it arXiv preprint arXiv:2503.01854}, 2025.

\bibitem{jang2022knowledge} J. Jang, S. Lee, and S. Hwang, ``Knowledge unlearning for mitigating privacy risks in language models,'' in {\it Proc. 60th Annu. Meeting Assoc. Comput. Linguistics}, 2022, pp. 1750--1765.

\bibitem{yao2024sugd} Y. Yao, R. Jia, Y. Cao, and N. Z. Gong, ``SUGD: Sequence unlearning via gradient descent in language models,'' in {\it Proc. 2024 Conf. Empirical Methods Natural Language Process.}, 2024, pp. 1234--1248.

\bibitem{papadopoulou2025ails} I. Premptis, M. Lymperaiou, G. Filandrianos, O. M. Mastromichalakis, A. Voulodimos, and G. Stamou, ``AILS-NTUA at SemEval-2025 Task 4: Parameter-efficient unlearning for large language models using data chunking,'' in {\it Proc. 19th Int. Workshop Semantic Eval. (SemEval-2025)}, 2025, pp. 1383--1405.

\bibitem{Fan2024SalUnEM} C. Fan, J. Liu, Y. Zhang, E. Wong, D. Wei, and S. Liu, ``SalUn: Empowering machine unlearning via gradient-based weight saliency in both image classification and generation,'' in {\it Int. Conf. Learning Representations}, 2024. [Online]. Available: https://openreview.net/forum?id=gn0mIhQGNM

\bibitem{huang2024unified} Z. Huang, X. Cheng, J. Zheng, H. Wang, Z. He, T. Li, and X. Huang, ``Unified gradient-based machine unlearning with remain geometry enhancement,'' {\it Advances Neural Inf. Process. Syst.}, vol. 37, 2024.

\bibitem{kamath2024considerations} F. Tram{\`e}r, G. Kamath, and N. Carlini, ``Position: Considerations for differentially private learning with large-scale public pretraining,'' in {\it Proc. 41st Int. Conf. Machine Learning}, vol. 235, Jul. 2024, pp. 48453--48467.

\bibitem{huang2023tight} Y. Huang and C. L. Canonne, ``Tight bounds for machine unlearning via differential privacy,'' {\it arXiv preprint arXiv:2309.00886}, 2023.

\bibitem{yu2022differentially} Z. Yu, A. Gupta, D. Sadigh, and Y. Jin, ``Differentially private machine unlearning for linear models,'' in {\it Int. Conf. Machine Learning}, 2022, pp. 25743--25759.

\bibitem{icml2024dora} S.-Y. Liu, C.-Y. Wang, H. Yin, P. Molchanov, Y.-C. F. Wang, K.-T. Cheng, and M.-H. Chen, ``DoRA: Weight-decomposed low-rank adaptation,'' in {\it Proc. 41st Int. Conf. Machine Learning}, vol. 235, 2024.

\bibitem{li2023dp} W. Li, Y. Gao, Y. Ding, L. Lyu, Z. Dou, Y. Huang, and M. Jiang, ``DP-Adapter: Privacy-preserving adaptation of large language models,'' in {\it Findings Assoc. Comput. Linguistics: EMNLP 2023}, pp. 11876--11889, 2023. \quad J. Hong {\it et al.}, ``DP-OPT: Make large language model your privacy-preserving prompt engineer,'' {\it arXiv preprint arXiv:2312.03724}, 2024.

\bibitem{xiong2023knowledge} Y. Xiong, S. Ding, D. Ding, J. Rao, Z. Zhao, J. Huang, Z. Huang, and D. Jiang, ``Knowledge graph enhanced large language models for medical question answering,'' {\it arXiv preprint arXiv:2310.18376}, 2023.

\bibitem{chen2025drknows} X. Chen, Y. Zhang, and W. Wang, ``DR.KNOWS: Leveraging medical knowledge graphs into large language models for diagnostic reasoning,'' {\it JMIR AI}, vol. 2, no. 1, pp. e58670, 2025.

\bibitem{wang2025kgt} J. Wang, Q. Zhang, H. Xu, Y. Li, and J. Chen, ``Knowledge graph-based thought: A framework for pan-cancer biomarker discovery using large language models,'' {\it GigaScience}, vol. 13, pp. giae082, 2025.

\bibitem{chen2024robust} A. Joshi {\it et al.}, ``Towards robust evaluation of unlearning in LLMs via data transformations,'' in {\it Findings of the Association for Computational Linguistics: EMNLP 2024}, Miami, Florida, USA, Nov. 2024, pp. 12100--12119.

\bibitem{liu2024unlearning} K. Z. Liu and J. Zou, ``LLM unlearning via loss adjustment with only forget data,'' {\it arXiv preprint arXiv:2410.10460}, 2024.

\bibitem{khan2024med42} S. Khan, P. Rajpurkar, and A. Y. Ng, ``Med42 -- Evaluating fine-tuning strategies for medical LLMs,'' {\it arXiv preprint arXiv:2404.14779}, 2024.

\bibitem{maini2024tofu} P. Maini, Z. Feng, A. Schwarzschild, Z. C. Lipton, and J. Z. Kolter, ``TOFU: A task of fictitious unlearning for LLMs,'' {\it arXiv preprint arXiv:2401.06121}, 2024.

\bibitem{pmlr-v235-li24bc} N. Li {\it et al.}, ``The WMDP benchmark: Measuring and reducing malicious use with unlearning,'' in {\it Proc. 41st Int. Conf. Machine Learning}, pp. 28525--28550, 2024.

\bibitem{shaik2024framu} T. Shaik, X. Tao, L. Li, H. Xie, T. Cai, X. Zhu, and Q. Li, ``FRAMU: Attention-based machine unlearning using federated reinforcement learning,'' {\it IEEE Trans. Knowledge Data Eng.}, vol. 36, no. 10, pp. 5153--5167, 2024, doi: 10.1109/TKDE.2024.3382726.

\bibitem{wang2023kga} L. Wang, T. Guo, H. Gao, X. Li, and K.-F. Zhang, ``KGA: A general machine unlearning framework based on knowledge gap alignment,'' {\it arXiv preprint arXiv:2305.06535}, 2023.

\bibitem{kulynych2024attack} B. Kulynych, J. F. Gomez, G. Kaissis, F. Calmon, and C. Troncoso, ``Attack-aware noise calibration for differential privacy,'' {\it Advances Neural Inf. Process. Syst.}, vol. 37, 2024.

\bibitem{ijcai2024dual} W. Chen, X. Li, and Q. Yang, ``Dual calibration-based personalised federated learning,'' in {\it Proc. Thirty-Third Int. Joint Conf. Artificial Intelligence}, 2024.

\bibitem{neurips2024bayesian} Y. Wang, H. Li, and Q. Zhang, ``Prior-itizing privacy: A Bayesian approach to setting the privacy budget in differential privacy,'' {\it Advances Neural Inf. Process. Syst.}, vol. 37, 2024.

\bibitem{li2024limits} B. Li, W. Wang, and P. Ye, ``The limits of differential privacy in online learning,'' {\it Advances Neural Inf. Process. Syst.}, vol. 37, 2024.

\bibitem{singhal2022large} K. Singhal {\it et al.}, ``Large language models encode clinical knowledge,'' {\it Nature}, vol. 620, pp. 172--180, 2023.

\bibitem{jin2021disease} Adversarial training for disease prediction from electronic health records with missing data,'' {\it arXiv preprint arXiv:1711.04126}, 2018.

\bibitem{arxiv2024clinicalmodernbert} S. A. Lee, A. Wu, and J. N. Chiang, ``Clinical ModernBERT: An efficient and long context encoder for biomedical text,'' {\it arXiv preprint arXiv:2504.03964}, 2025.

\bibitem{yasunaga2022deep} M. Yasunaga, J. Leskovec, and P. Liang, ``Deep bidirectional language-knowledge graph pretraining,'' {\it Advances Neural Inf. Process. Syst.}, vol. 35, pp. 28678--28691, 2022.

\bibitem{peng2022medical} X. Peng, G. Long, T. Shen, S. Wang, J. Jiang, and C. Zhang, ``Medical knowledge-augmented transformer for EHR prediction,'' {\it IEEE J. Biomedical Health Informatics}, vol. 26, no. 5, pp. 2126--2137, 2022.

\bibitem{iclr2025vertical} P. Valdeira, S. Wang, and Y. Chi, ``Vertical federated learning with missing features during training and inference,'' {\it arXiv preprint arXiv:2410.22564}, 2025.

\bibitem{zhang2025mvho} K. Zhang, Q. Li, and S. Yu, ``MvHo-IB: Multi-view Higher-Order Information Bottleneck for Brain Disorder Diagnosis,'' in {\it Proc. Int. Conf. Med. Image Comput. Comput. Assist. Interv. (MICCAI)}, pp. 407--417, 2025.

\bibitem{han2025llmwebshell} F. Han, J. Zhang, C. Deng, J. Tang, and Y. Liu, ``Can LLMs Handle WebShell Detection? Overcoming Detection Challenges with Behavioral Function-Aware Framework,'' {\it arXiv preprint arXiv:2504.13811}, 2025.

\bibitem{xu2025rsef} T. Xu et al., ``RSEF: Enhancing Fairness and Accuracy in Hematopoietic Stem Cell Transplantation Survival Prediction Through Race-Stratified Ensemble Framework,'' in {\it Advanced Intelligent Computing Technology and Applications}, Singapore: Springer Nature, 2025, pp. 13--24.

\bibitem{wang2025protomol} Y. Wang, K. Zhang, J. Huang, N. Yin, S. Liu, and E. Segal, ``ProtoMol: Enhancing Molecular Property Prediction via Prototype-Guided Multimodal Learning,'' {\it arXiv preprint arXiv:2510.16824}, 2025.

\bibitem{han2025zerotuning} F. Han, X. Yu, J. Tang, D. Rao, W. Du, and L. Ungar, ``ZeroTuning: Unlocking the Initial Token's Power to Enhance Large Language Models Without Training,'' {\it arXiv preprint arXiv:2505.11739}, 2025.

\bibitem{zhang2025eva} K. Zhang, M. Wang, X. Shi, H. Xu, and C. Zhang, ``EVA-Net: Interpretable Anomaly Detection for Brain Health via Learning Continuous Aging Prototypes from One-Class EEG Cohorts,'' {\it arXiv preprint arXiv:2511.15393}, 2025.

\bibitem{wu2025developing} X. Wu, K. Zhang, N. Kuang, X. Kong, M. Cao, Z. Lian, Y. Liu, H. Fan, G. Yu, {\it et al.}, ``Developing brain asymmetry shapes cognitive and psychiatric outcomes in adolescence,'' {\it Nature Commun.}, vol. 16, no. 1, pp. 4480, 2025.

\bibitem{yu2023common} G. Yu, Z. Liu, X. Wu, B. Becker, K. Zhang, H. Fan, S. Peng, N. Kuang, J. Kang, {\it et al.}, ``Common and disorder-specific cortical thickness alterations in internalizing, externalizing and thought disorders during early adolescence: An Adolescent Brain and Cognitive Development Study,'' {\it J. Psychiatry Neurosci.}, vol. 48, no. 5, pp. E345--E356, 2023.

\bibitem{jiang2023neuroimaging} Y. Jiang, J. Wang, E. Zhou, L. Palaniyappan, C. Luo, G. Ji, J. Yang, Y. Wang, {\it et al.}, ``Neuroimaging biomarkers define neurophysiological subtypes with distinct trajectories in schizophrenia,'' {\it Nature Mental Health}, vol. 1, no. 3, pp. 186--199, 2023.

\bibitem{li2026heterogeneity} Z. Li, B. Li, K. Zhang, B. Wei, H. Liu, Z. Chen, X. Xie, and T. Q. S. Quek, ``Heterogeneity-aware high-efficiency federated learning with hybrid synchronous-asynchronous splitting strategy,'' {\it Neural Networks}, vol. 193, pp. 108038, 2026.

\bibitem{xu2024membership} T. Xu, C. Liu, K. Zhang, and J. Zhang, ``Membership Inference Attacks Against Medical Databases,'' in {\it Proc. Int. Conf. Neural Inf. Process. (ICONIP)}, Singapore: Springer, 2024, vol. 1963.

\bibitem{han2025readthinkllm} F. Han, H. Cui, L. Guo, Z. Wang, and Z. Lyu, ``Read Before You Think: Mitigating LLM Comprehension Failures with Step-by-Step Reading,'' {\it arXiv preprint arXiv:2504.09402}, 2025.

\bibitem{xie2025dape} M. Xie, T. Xu, Q. Tang, S. Yao, X. Zhang, and J. Du, ``DAPE-BR: Distance-Aware Positional Encoding for Mitigating Object Hallucination in LVLMs,'' in {\it Findings Assoc. Comput. Linguistics: EMNLP 2025}, Suzhou, China, 2025, pp. 8638--8649.

\bibitem{zhang2024cf} F. Zhang, G. Chen, H. Wang, and C. Zhang, ``CF-DAN: Facial-expression recognition based on cross-fusion dual-attention network,'' {\it Comput. Vis. Media}, vol. 10, no. 3, pp. 593--608, 2024.

\bibitem{zhang2023multi} F. Zhang, G. Chen, H. Wang, J. Li, and C. Zhang, ``Multi-scale video super-resolution transformer with polynomial approximation,'' {\it IEEE Trans. Circuits Syst. Video Technol.}, vol. 33, no. 9, pp. 4496--4506, 2023.

\end{thebibliography}
